\documentclass{article}
\usepackage[numbers,sort]{natbib}
\usepackage[final]{nips_2017}

\usepackage[utf8]{inputenc} 
\usepackage[T1]{fontenc}    
\usepackage{hyperref}       
\usepackage{url}            
\usepackage{graphicx}
\usepackage{booktabs}       
\usepackage{amsfonts}       
\usepackage{amssymb}
\usepackage{amsmath}
\usepackage{amsthm}
\usepackage{algorithm}
\usepackage{algorithmic}
\usepackage{nicefrac}       
\usepackage{microtype}      
\usepackage{subfig}
\title{Interactive Reinforcement Learning for Object Grounding via Self-Talking}

\author{\quad\quad\quad Yan Zhu$^1$\thanks{Yan Zhu works at Facebook Applied Machine Learning now. The majority of the work was done at Rutgers.}\quad\quad\quad Shaoting Zhang$^2$\quad\quad\quad Dimitris Metaxas$^1$\\
    \small{$^1$Rutgers University\quad\quad $^2$Baidu Research}\\
    \small{$^1$\texttt{\{yz328, dnm\}@cs.rutgers.edu}\quad\quad$^2$\texttt{zhangshaoting@baidu.com}}
}

\newcommand{\app}{\raise.17ex\hbox{$\scriptstyle\sim$}}

\newcommand{\tss}[1]{\textsuperscript{#1}}

\def\etal{\emph{et al}}
\newcommand{\myparagraph}[1]{\smallskip\noindent\textbf{#1}}

\begin{document}

\maketitle

\begin{abstract}
Humans are able to identify a referred visual object in a complex scene via a few rounds of natural language communications. Success communication requires both parties to engage and learn to adapt for each other.

In this paper, we introduce an interactive training method to improve the natural language conversation system for a visual grounding task. During interactive training, both agents are reinforced by the guidance from a common reward function. The parametrized reward function also cooperatively updates itself via interactions, and contribute to accomplishing the task. We evaluate the method on GuessWhat?! visual grounding task, and significantly improve the task success rate. However, we observe language drifting problem during training and propose to use reward engineering to improve the interpretability for the generated conversations. Our result also indicates evaluating goal-ended visual conversation tasks require semantic relevant metrics beyond task success rate.
\end{abstract}

\section{Introduction}\label{sec:chap3_intro}

Natural language interaction is probably the most efficient and natural way for humans to acquire knowledge and exchange information. In artificial intelligence, developing intelligent agents that can correspond the linguistic concepts with the visual sensor inputs, and communicate via goal-ended dialogue is also a fundamental problem.

Recently, two visual grounded conversation tasks along with datasets/environments are proposed: VisDial \cite{das2016visual} and GuessWhat?! \cite{de2016guesswhat}. VisDial \cite{das2016visual} collects a large dataset containing free-form conversations about natural images. Based on VisDial dataset, Das \etal  \cite{das2017learning} further proposed an \textit{image level} grounding task and applied reinforcement learning to train a goal-ended conversation system. The generated conversation was mainly evaluated on an image retrieval task.

Instead, GuessWhat?! proposed an environment focusing on \textit{object instance level} grounding. 
The object grounding task implicitly requires agents detect and recognize object instances, understand the spatial layout, and use positional / attribute word to reason and distinguish among candidate regions. In this paper, we focus on GuessWhat?! task, since object level grounding is more practically relevant to real-world applications, for instance, interactive navigation robots.

Beyond supervised training using human dialog, reinforcement learning has recently been adopted in visual conversation \cite{das2017learning, strub2017end}. Particularly, on GuessWhat?! task, Strub \etal \cite{strub2017end} used RL to tune the question generator while keeping answer agent and guesser unchanged. However, human collaboration normally requires both parties to engage and adapt for the other. In this paper, we propose to interactive train all three models in a dynamic environment. The question generator and the answer models are collectively tuned by a common reward function using reinforcement learning. The reward function, which is parameterized by the guesser model, is also dynamically updated to cooperate with other two models. Our result significantly outperforms the previous best result on GuessWhat?! task and achieves near human performance. Despite improved task success rate, we observe the generated conversations suffer from language drifting problem. We also propose a reward engineering technique to help improve interpretability of the generated conversations. 

Our main result shows that agents are able to achieve near human level performance on visual object grounding task, by drifting from natural language towards a contrived language. It also indicates current goal-ended visual conversation task requires more semantic related metrics for evaluation, other than task completion rate. 
In summary, the contribution of this paper is two-fold:
\vspace{-2mm}
\begin{itemize}
\item We introduce an interactive training method for object instance grounding task. The proposed training method significantly outperforms previous best results on GuessWhat?! task. 

\item To balance between interpretability and task success rate, we propose a reward engineering technique to interfere training, and improves the readability of the generated conversations.
\end{itemize}

\section{Related Work}\label{sec:chap3_related}
\vspace{-2mm}

\myparagraph{Visual Conversation:} Existing visual conversation datasets can be divided into task-oriented dialogue \cite{de2016guesswhat} and free-form (chit-chat) dialogue \cite{das2016visual}. Recently, Das \etal \cite{das2017learning} also introduced a goal-ended conversation task based on VisDial. One salient difference is that GuessWhat?! focuses on \textit{object instance level} grounding while the task in \cite{das2017learning} focus on \textit{image level} grounding. Beyond supervised training baselines, Das \etal. \cite{das2017learning} showed cooperative RL improves supervised trained baselines in VisDial image retrieval task. Also on GuessWhat?!, Strub \etal.  \cite{strub2017end} also showed that RL could improve task success rate by only tuning the question generator and keeping other two models static. Notably, the reward function in GuessWhat?! task depends on the \textit{subjective} parametric guesser model rather than an objective metric. In this work, we focus on the object grounding task in GuessWhat?!, and extend reinforcement training towards a more interactive setting: both conversation bots are collectively trained using RL, and the parameterized reward function (the guesser model) is also actively involved in the updating dynamics.

\myparagraph{Artificial Language vs Natural Language:} Recently, \cite{mordatch2017emergence, evtimova2017emergent, das2017learning} found that multiple agents can develop their own communication protocols (artificial languages) during cooperative training. The emerged language is very effective between AI agents, but not interpretable for humans \cite{Chattopadhyay2007}. To narrow the semantic gap between artificial language and natural language, \cite{Satwik2007, Milli2007} explored different techniques to retain interpretability, including constrain vocabulary size \cite{Satwik2007} and iteratively updating different agents \cite{Milli2007}. These findings are based on synthetic environments, rather than natural image based tasks. To echo these findings, we observe similar language drifting problems during interactive training. To improve the interpretability, we choose to explicitly enforce the desired dialogue properties in the reward function, thus balancing the trade-off between task success rate and interpretability.

\section{Model Architecture and Interactive Training}\label{sec:chap3_method}
\vspace{-2mm}
We generally follow the architecture design in \cite{de2016guesswhat} for answer model and guesser. The only salient architecture difference is that we use seq2seq model for question generator instead of vanilla LSTMs.

\myparagraph{seq2seq Question Generator}\quad The seq2seq model was originally introduced for machine translation, and later adopted in conversation systems. Compared with vanilla LSTM, seq2seq can be extended with attention modules for long distance reasoning. We used a global dot product attention layer to combine the visual context with the language embeddings. 

The seq2seq model first encodes the previous conversation history using the LSTM encoder, then the language embeddings are mixed with the image feature (we also use VGG features) in the attention module. At reinforcement training stage, we only take at most 2 round of recent conversations as the input to the seq2seq model, instead of the whole conversation history.

\myparagraph{Interactive Reinforcement Training}
\quad After the supervised pre-training, three models only obtain knowledge from the static dataset, but not yet learn to cooperate with each other via interaction. 

We argue that success communication requires \textit{all} parties learn to adapt for others. Based on this intuition, we enable three models to actively learn in a self-talking environment in an interactive manner.  

In \cite{de2016guesswhat}, the reward function for the question generator is a binary score, dependent on whether the guesser finish the task: so the reward at round $t$ is $R_t(\mathbf{s}_t, (q_t, a_t):\theta_{g}) \in \{0, 1\}$, where $\theta_{g}$ is guesser's parameter, $\mathbf{s}_t$ is the state of question generator and answer model. We use the same reward definition to update the answer model. For the answer model, we also augment a score branch (a single FC + ReLU layer), to estimate the reward value, in order to stabilize the policy gradient updates. 

With the above extension, we want to maximize the expected reward over two models's policies, parameterized by $\theta_{a}$ and $\theta_{q}$: $ J(\theta_{a}, \theta_{q}) = \mathop{\mathbb{E}}\limits_{\pi_{a}, \pi_{q}} \lbrack \Sigma_{t=0}^{T} R_t(\mathbf{s}_t, (q_t, a_t))\rbrack$. \quad The policy gradient updates for the question generator and the answer model can be written as follows:
\vspace{-1mm}
\small
\begin{equation}
\mathop{\nabla}\limits_{\theta_{q}} J = \mathop{\mathbb{E}}\limits_{\pi_{a}, \pi_{A}} [\sum_{t=1}^{T} \mathop{\nabla}\limits_{\theta_{q}} log \pi_{q}(\mathbf{q_t} | \mathbf{s}^{t}_{q}) * (r- b_{q}))];   \quad \quad 
\mathop{\nabla}\limits_{\theta_{a}} J = \mathop{\mathbb{E}}\limits_{\pi_{q}, \pi_{a}} [\sum_{t=1}^{T}\mathop{\nabla}\limits_{\theta_{a}} log \pi_{a}(\mathbf{a_t}  |\mathbf{s}^{t}_{a}) * (r- b_{a}))]
\vspace{-2mm}
\label{eq:chap3_grad_qgen}
\end{equation}
\normalsize
\vspace{-1mm}

Note that the evaluation of reward $R$ depends on the guesser model. The accuracy of the guesser model is indeed far from perfect even for human dialogue ($30\%$ errors). The mismatch between generated conversation and human conversation further enlarges the error and affects the policy gradient. Therefore, we let the guesser tune itself on the generated dialogue. Guesser's parameter $\theta_{g}$ is updated by optimizing cross entropy loss of guesser's prediction using generated conversations:
\small
\begin{equation}
\mathop{\nabla}\limits_{\theta_{g}} \mathit{L}_{CE}(\mathbf{O}_{gt}, Guesser((q,a)_{:T};\theta_{g}))
\label{eq:chap3_grad_guesser}
\end{equation}
\normalsize
\vspace{-2mm}

\vspace{-3mm}
\myparagraph{Reward Engineering} \quad \quad Above interactive learning effectively improves task success rate, but the generated conversations diverge from natural language towards an effective but uninterpretable communication protocols. One explanation is that during interactive training, the guesser model manages to tolerate the gradually shifted conversation, and feedback positive reward ``over-generously''. 

Based on the above assumption, we use several heuristics to prune unnatural generated questions before feeding the generated conversation into the guesser. The intention is to limit the guesser only read the ``natural'' part of the conversation, and explicitly discourage the guesser to squeeze signal from unnatural QAs. Specifically, we use two heuristics to prune the unnatural QAs: 1) Removing questions containing repetitive words/phrases (e.g. ``is it in front left front left front left?''); 2) Removing near duplicate questions happened in earlier conversations (e.g. ``Is it on the left? ... On the left?''). In an extreme case, if the generated QAs are mostly unreadable and pruned, the guesser won't get enough input and forced to fail, so that the reward will be zero. 

This way, we explicitly inject our preference for the generated text. As a result, we effectively interfere the rewarding function, thus it can be viewed as a form of reward engineering \cite{deway2014}.

\section{Experiments} \label{sec:chap3_exp}

\vspace{-2mm}
\myparagraph{Implementation details}  \quad We use similar data preprocessing as \cite{strub2017end}. We also do supervised pre-training and get comparable error rate as in \cite{strub2017end} (answer model: 0.213, guesser: 0.380, question generator: 0.581). At interactive learning stage, we use Adam optimizer with batch size 64 with learning rate 1e-4. To generate questions, we use multinomial sampling in both training and testing.
 
\begin{figure}
\centering\footnotesize

\subfloat[Evaluation: task success rate (SR) and quality of questions and answers  on test set.\label{table:chap3_main}]{
 \addtolength{\tabcolsep}{2pt}
 \centering\small\renewcommand\arraystretch{1.5}
\begin{tabular}{@{\hskip 0mm}l|ccc|cccccc@{\hskip 0mm}}
 & SL  & RL\cite{strub2017end} &  Human & RL\tss{Q}  & IRL\tss{QA} & IRL\tss{QG} &IRL\tss{AG} & IRL\tss{QAG} &IRL-prune\tss{QAG} \\
\hline
SR & .417  & .603 & .844 &.582 &.651 &.631 &.777 &  \textbf{.829} &.813\\
Q-que & - & - & - &\textbf{1.71} &1.97 & 2.71 & - & 4.98 &3.53\\
Q-ans &.780 &  &- &\textbf{.915} & .801 &.909 &.408 &.590 &.681

\end{tabular}
}
\vfill
\subfloat[Examples of generated conversation from IRL-prune\tss{QAG} (top right in each cell) and SL baseline model (bottom right). The ground truth target region is highlighted by blue bounding boxes. \label{table:chap3_good_examples}]{
\includegraphics[width=0.50\textwidth]{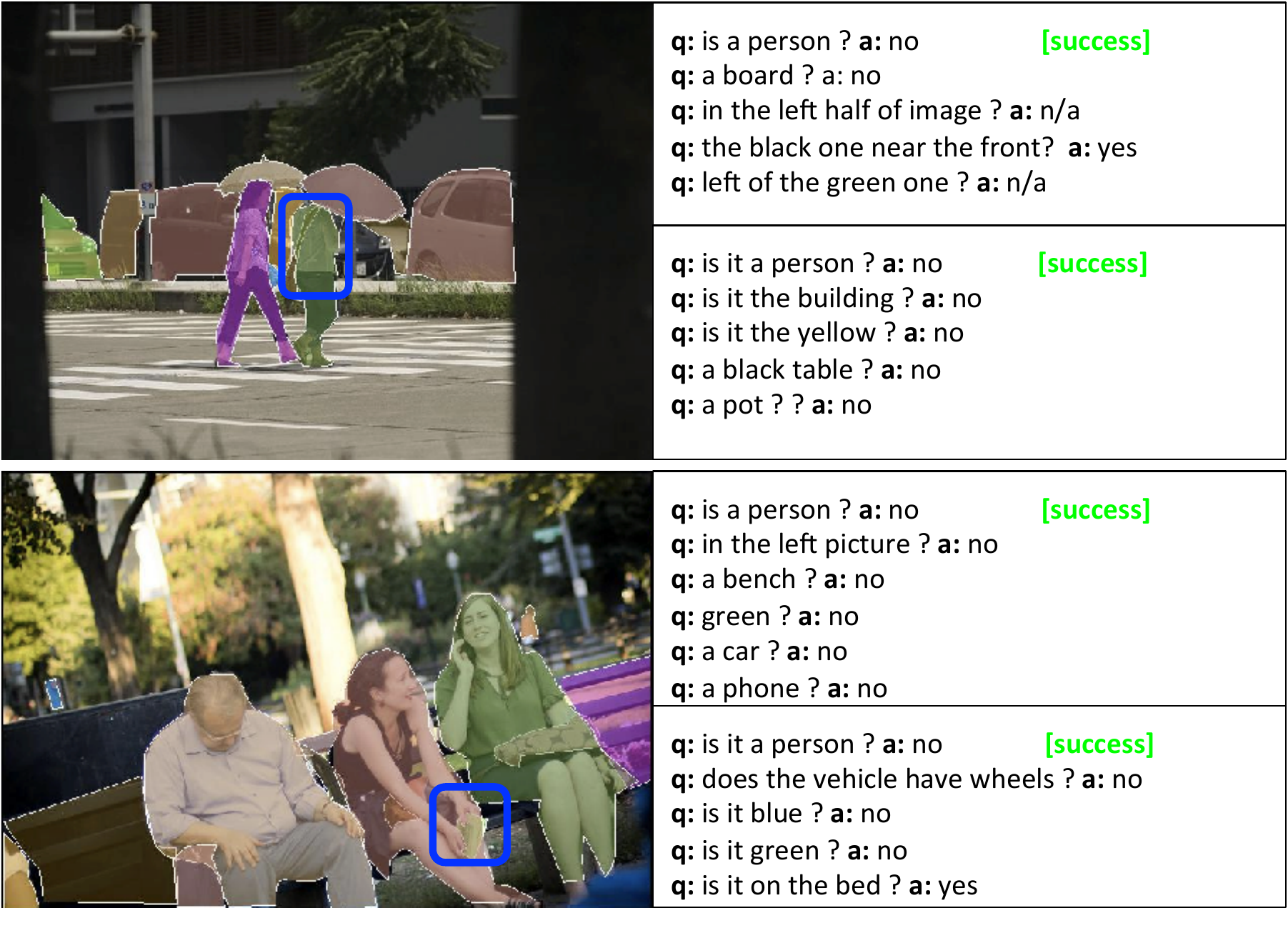} \quad
\includegraphics[width=0.495\textwidth]{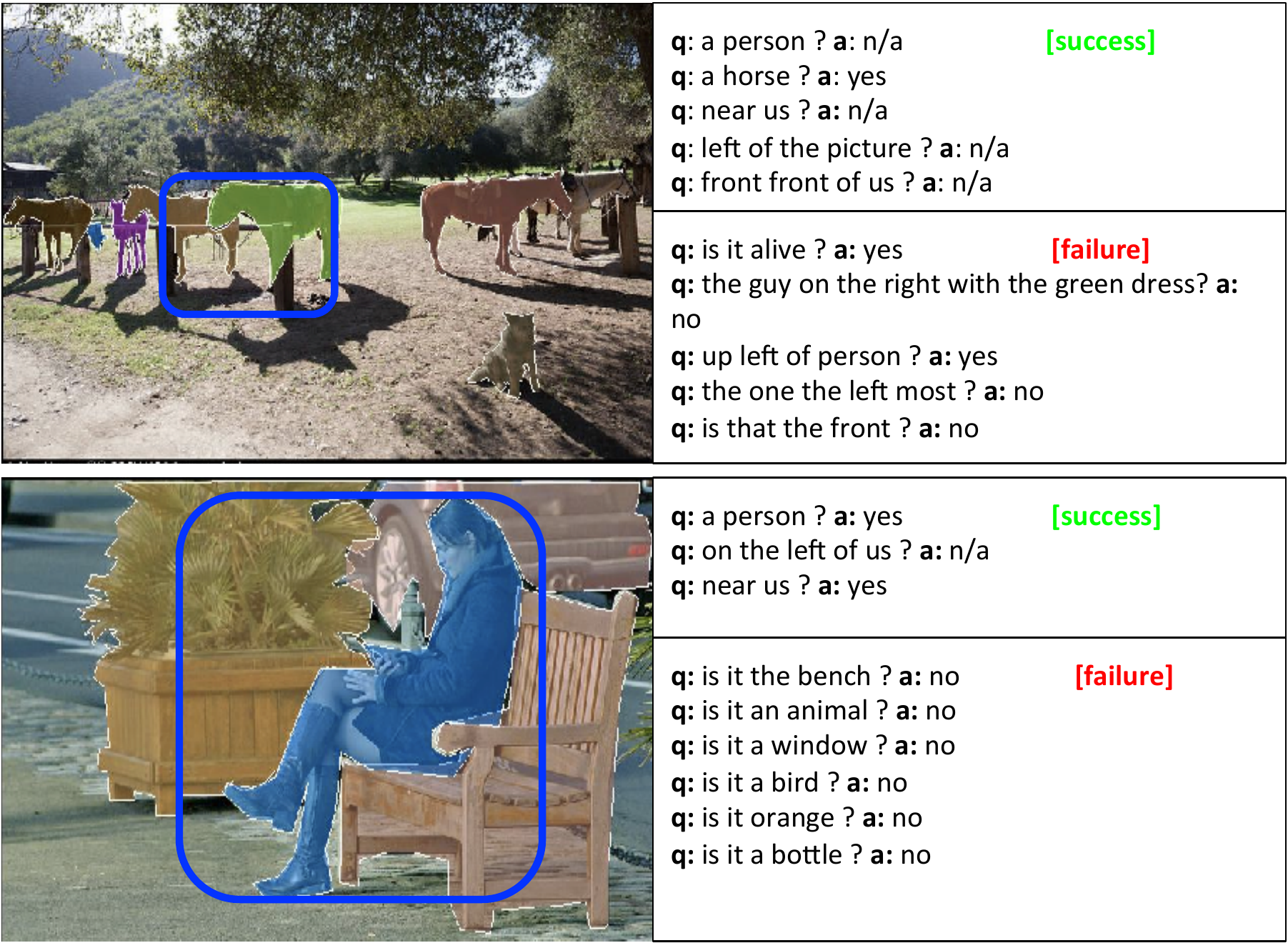}
}
\end{figure}

\myparagraph{Task Success Rate:} The task success rates for baseline models and different variants of interactive trained models are shown in Table \ref{table:chap3_main}. In table \ref{table:chap3_main}, we use IRL to denote interactive reinforcement learning and use the superscript to denote which models are actively tuned during training. All interactively trained models consistently outperform the baseline RL model \cite{strub2017end}. \footnote{The best score for baseline \cite{strub2017end} is copied from the author's Github page} As expected, the most successful model is IRL\tss{QAG}, which is very close to the human score as measured in \cite{de2016guesswhat}. 

\myparagraph{Quantifying Semantic Gap:} Although effectively improve task success rate, IRL\tss{QAG} tends to generate uninterpretable conversations. Typical example are 1) repetitive word/phrase in questions; 2)  the answer model tends to use ``n/a'' more frequently, even if the rational answers are ``yes'' or ``no''. (top left example in Figure \ref{table:chap3_good_examples}).
We setup human studies to quantify this semantic gap: 
First, we asked independent human subjects to evaluate the generated \textit{answer quality}. We randomly generated 100 cases and asked 20 human subjects to decide whether the generated answers agree with their judgments. Each QA pair is evaluated by 3 subjects with binary scores (3rd row in Table \ref{table:chap3_main}).  When guesser and answer model are jointly updated (IRL\tss{AG}), generated answers tend to disagree with humans. However, when the question generator is jointly updated, the answer quality improves, probably because generator adapts its questions to make it easy for answer model.

Second, we asked human subjects to evaluate \textit{question quality} in terms of 1) whether the question is interpretable and 2) relevant to the image content. We asked the subjects to rank the quality of the generated questions from different models (best rank is 1). The averaged ranking is shown in the 2nd row of Table \ref{table:chap3_main}. 
IRL\tss{QAG}'s semantic gap is enlarged most, despite improving success rate.

The reward engineering version IRL-prune\tss{QAG} improves interpretability compared with IRL\tss{QAG}, but the task success rate is also slightly degraded. Some qualitative examples shown in Figure \ref{table:chap3_good_examples}.

\vspace{-3mm}
\section{Conclusion} \label{sec:chap3_conclude}
\vspace{-2mm}
We proposed an interactive training method on object instance level visual grounding conversation task and significantly improve task success rate. Observing the language drifting problem during the interactive learning, we proposed a reward engineering technique during training and improved interpretability. The major problem of our method is still language drifting. Our result also suggests visual goal-ended conversation need semantic evaluation metric other than task success rate.
\setlength{\bibsep}{0pt plus 0.ex}
\bibliographystyle{plainnat}
{
\footnotesize
\bibliography{mybib}
}

\end{document}